\documentclass[10pt,twocolumn,letterpaper]{article}

\usepackage{iccv}
\usepackage{times}
\usepackage{epsfig}
\usepackage{graphicx}
\usepackage{amsmath}
\usepackage{amssymb}
\usepackage{algorithm}
\usepackage{subfigure}
\usepackage{bm,amsfonts,amssymb}
\usepackage{times}
\usepackage{multirow,array}
\usepackage{color}
\usepackage{epstopdf}
\usepackage{url,color}

\usepackage[pagebackref=true,breaklinks=true,letterpaper=true,colorlinks,bookmarks=false]{hyperref}
\iccvfinalcopy 

\def\iccvPaperID{2040} 

\ificcvfinal\fi
\begin{document}

\title{Mixed High-Order Attention Network for Person Re-Identification}

\author{Binghui Chen, Weihong Deng\thanks{Corresponding author}, Jiani Hu\\
Beijing University of Posts and Telecommunications\\
{\tt\small chenbinghui@bupt.edu.cn, whdeng@bupt.edu.cn, jnhu@bupt.edu.cn}
}

\maketitle

\begin{abstract}
   Attention has become more attractive in person re-identification (ReID) as it is capable of biasing the allocation of available resources towards the most informative parts of an input signal. However, state-of-the-art works concentrate only on coarse or first-order attention design, e.g. spatial and channels attention, while rarely exploring higher-order attention mechanism. We take a step towards addressing this problem. In this paper, we first propose the High-Order Attention (HOA) module to model and utilize the complex and high-order statistics information in attention mechanism, so as to capture the subtle differences among pedestrians and to produce the discriminative attention proposals. Then, rethinking person ReID as a zero-shot learning problem, we propose the Mixed High-Order Attention Network (MHN) to further enhance the discrimination and richness of attention knowledge in an explicit manner.  Extensive experiments have been conducted to validate the superiority of our MHN for person ReID over a wide variety of state-of-the-art methods on three large-scale datasets, including Market-1501, DukeMTMC-ReID and CUHK03-NP.
   \textcolor[rgb]{1,0,0}{Code is available at http://www.bhchen.cn/}.
\end{abstract}

\section{Introduction}
Since the quest for algorithms that enable cognitive abilities is an important part of machine learning, person re-identification (ReID) has become more attractive, where the model is requested to be capable of correctly matching images of pedestrians across videos captured from different cameras. This task has drawn increasing attention in many computer vision applications, such as surveillance \cite{wang2013intelligent}, activity analysis \cite{loy2009multi,loy2010time} and people tracking \cite{yu2013harry,tang2017multiple}. It is also challenging because the images of pedestrians are captured from disjoint views, the lighting-conditions/person-poses differ across cameras, and occlusions are frequent in real-world scenarios.

\begin{figure}
\vspace{-2em}
  \centering
  \includegraphics[width=0.8\linewidth]{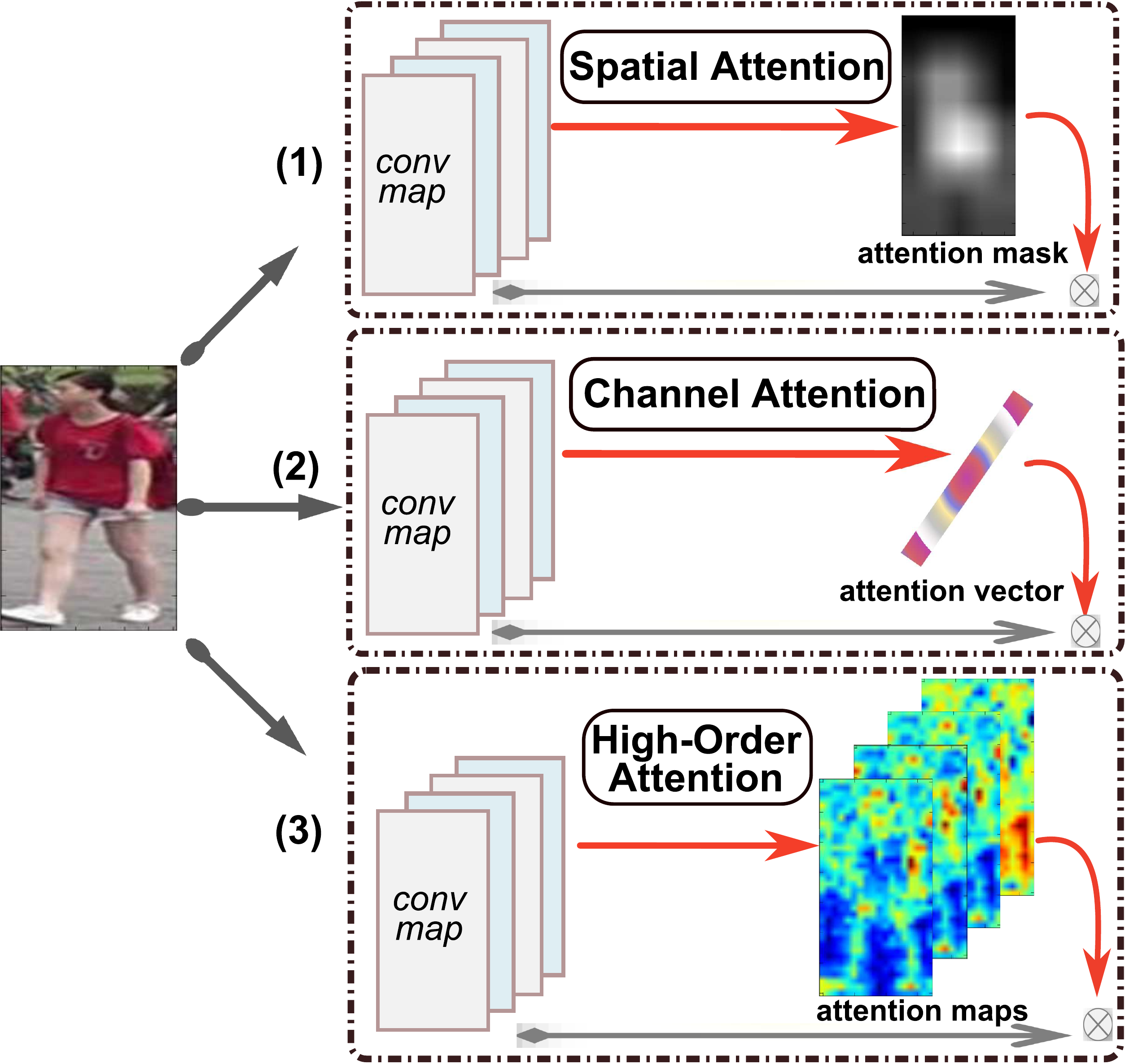}\\
  \caption{Attention comparison. (1) \emph{Spatial attention} uses softmax-like gated functions to produce a spatial mask. (2) \emph{Channel attention} \cite{hu2018squeeze} uses global average pooling and fully connected layers to produce a scale vector. (3) Our \emph{high-order attention} uses high-order polynomial predictor to produce scale maps that contain high-order statistics of convolutional activations.}\label{fig_sp_ca_ho_attention}
  \vspace{-2em}
\end{figure}

Affected by the aforementioned factors, the discrimination of feature representations of pedestrian images actually is not good enough. In order to obtain discriminative feature representations, many research works \cite{liu2017hydraplus,li2018harmonious,li2018diversity,xu2018attention,kalayeh2018human,zhao2017spindle,varior2016gated} resort to attention mechanism so as to highlight the informative parts (e.g. spatial locations) of convolutional responses and suppress the noisy patterns (e.g. background). Specifically, \emph{spatial attention} \cite{li2018diversity,li2018harmonious,xu2018attention} is a form of visual attention that involves directing attention to a location in space, it allows CNN to selectively process visual information of an area within the visual field. While, in spatial attention, the processing strategy of spatial masking is coarse and has no intrinsic effect on modulating the fine-grained channel-knowledge. Recently, \emph{channel attention} \cite{chen2017sca,hu2018squeeze,li2018harmonious} is proposed to adaptively recalibrates channel-wise convolutional responses by explicitly modelling interdependencies among channels. And the combination of spatial and channel attention has also been successfully applied in person ReID \cite{li2018harmonious}. However, we emphasize that these commonly used attention methods (i.e. spatial and channel attention) are either coarse or first-order, being confined to mining only simple and coarse information, in person ReID cases, they are insufficiently rich to capture the complex/high-order interactions of visual parts and the subtle differences among pedestrians caused by various viewpoints/person poses, as a result, the produced attention maps are neither discriminative or detailed. To this end, we dedicate to modeling the attention mechanism via high-order statistics of convolutional activations so as to capture more complex and high-order relationships among parts and to produce powerful attention proposals.

Moreover, we rethink the problem of person ReID as a \emph{zero-shot} learning (ZSL) task where there is no intersection of pedestrian identities between training and testing sets. \emph{Zero-shot} learning has large gap with conventional \emph{full-shot} learning (e.g. classification on CIFAR \cite{chen2018virtual,Chen_2017_CVPR}, Imagenet \cite{russakovsky2015imagenet}), and in zero-shot settings, the phenomenon of `\emph{partial/biased learning behavior of deep model}' \cite{chen2019energy} largely affects the embedding performance, i.e. the deep model will only focus on the biased visual knowledges that only benefit to the \emph{seen} identities and ignore the other helpful ones that might be useful for identifying the \emph{unseen} identities. In other words, deep models easily learn to focus on surface statistical regularities rather than more general abstract concepts. However, many ReID works ignore this intrinsic problem of \emph{zero-shot} learning. To this end, proposing detail-preserving attention framework remains important.

In this paper, we first propose \emph{\textbf{High-Order Attention}} (HOA) module, a novel and powerful attention mechanism, to model the complex and high-order relationships among visual parts via high-order polynomial predictor, such that the subtle differences among pedestrian can be captured and discriminative attention results can be produced. Then, rethinking person ReID as a \emph{zero-shot} problem, we propose \emph{\textbf{Mixed High-Order Attention Network}} (MHN) to prevent the problem of `biased learning behavior of deep model` \cite{chen2019energy} and to encourage the richness of attention information. It is mainly achieved by employing multiple HOA modules with different orders to model diverse high-order statistics, such that all-sided attention knowledge can be preserved and thus the unseen pedestrian identity can be successfully recognized. Additionally, we introduce the adversarial learning constraint for MHN to further prevent the order collapse problem during training \footnote{Although, the proposed high-order attention module has ability to capture complex and high-order statistics, but suffering from `biased learning behavior of deep model`, in zero-shot settings, the high-order module might collapse to lower-order module.}, so as to explicitly enhance the discrimination of MHN. Our contributions can be summarized as follows:
\begin{itemize}
\vspace{-0.5em}
  \item The High-Order Attention (HOA) module is proposed to capture and use high-order attention distributions. To our knowledge, it is the first work to propose and apply high-order attention module in Person-ReID.
      \vspace{-0.5em}
  \item We rethink ReID as zero-shot learning task and propose the Mixed High-Order Attention Network (MHN) to efficiently utilize multiple HOA modules, so as to enhance the richness of attention by explicitly suppressing the `biased learning behavior of deep model`. And adversary learning constraint is introduced to further prevent the problem of order collapse.
      \vspace{-0.5em}
  \item MHN is a generally applicable and model-agnostic framework, it can be easily applied in the popular baseline architectures, such as IDE \cite{zheng2016person} and PCB \cite{sun2018beyond}.
      \vspace{-0.5em}
  \item Extensive experiments demonstrate the superiority of the proposed MHN over a wide range of state-of-the-art ReID models on three large benchmarks, i.e. Market-1501 \cite{zheng2015scalable}, DukeMTMC-ReID \cite{ristani2016performance,zheng2017unlabeled} and CUHK03-NP \cite{li2014deepreid,zhong2017re}.
      \vspace{-0.5em}
\end{itemize}
\section{Related work}
\textbf{Person ReID \& Attention Mechanism}: Person ReID intends to correctly match images of pedestrians across videos captured from different cameras, it has been widely studied, such as ranking by pairwise constraints \cite{paisitkriangkrai2015learning,wang2014person}, metric learning \cite{yi2014deep,xiong2014person}, deep embedding learning \cite{zheng2016person,sun2018beyond}, re-ranking \cite{zheng2015query,garcia2015person} and attributes learning \cite{su2016deep,zhao2014learning}. Recently, attention methods \cite{chen2017sca,xu2015show,hu2018squeeze,vaswani2017attention} in deep community are more attractive, in this paper, we focus on improving the performance of ReID via attention strategy.

Attention serves as a tool to bias the allocation of available resources towards the most informative parts of an input. Li et al. \cite{li2017learning} propose a part-aligning CNN network for locating latent regions (i.e. hard attention) and then extract and exploit these regional features for ReID. Zhao et at. \cite{zhao2017deeply} employ the Spatial Transformer Network \cite{jaderberg2015spatial} as the hard attention model for finding discriminative image parts. Except hard attention methods, soft attention strategies are also proposed to enhance the performance of ReID. For example, Li et at. \cite{li2018diversity} use multiple spatial attention modules (by softmax function) to extract features at different spatial locations. Xu et al. \cite{xu2018attention} propose to mask the convolutional maps via pose-guided attention module. Li et al. \cite{li2018harmonious} employ both the softmax-based spatial attention module and channel-wise attention module \cite{hu2018squeeze} to enhance the convolutional response maps. However, \emph{spatial attention} and \emph{channel attention} are coarse and first-order respectively, and are not capable of modeling the complex and high-order relationships among parts, resulting in loss of fine-grained information. Thus, to capture detailed and complex information, we propose High-Order Attention (HOA) module.

\textbf{High-order statistics}: It has been widely studied in traditional machine learning due to its powerful representation ability. And recently, the progresses of challenging fine-grained visual categorization task demonstrates integration of high-order pooling representations with deep CNNs can bring promising improvements. For example, Lin et al. \cite{lin2015bilinear} proposed bilinear pooling to aggregate the pairwise feature interactions. Gao et al. \cite{gao2016compact} proposed to approximate the second-order statistics via Tensor Sketch \cite{pham2013fast}. Yin et al. \cite{cui2017kernel} aggregated higher-order statistics by iteratively applying the Tensor Sketch compression to the features. Cai et al. \cite{cai2017higher} used high-order pooling to aggregate hierarchical convolutional responses. Moreover, the bilinear pooling and high-order pooling methods are also applied in Visual-Question-Answering task, such as \cite{fukui2016multimodal,kim2016hadamard,yu2017multi,yu2018beyond}. However, different from these above methods which mainly focus on using high-order statistics on top of feature pooling, resulting in high-dimensional feature representations that are not suitable for efficient/fast pedestrian search, we instead intend to enhance the feature discrimination by attention learning. We model high-order attention mechanism to capture the high-order and subtle differences among pedestrians, and to produce the discriminative attention proposals.

\textbf{Zero-Shot Learning}: In ZSL, the model is required to learn from the \emph{seen} classes and then to be capable of utilizing the learned knowledge to distinguish the \emph{unseen} classes. It has been studied in image classification \cite{li2018discriminative,changpinyo2016synthesized}, video recognition \cite{dalton2013zero} and image retrieval/clustering \cite{chen2019energy}. Interestingly, person ReID matches the setting of ZSL well where training identities have no intersection with testing identities, but most the existing ReID works ignore the problem of ZSL. To this end, we propose Mixed High-Order Attention Network (MHN) to explicitly depress the problem of `biased learning behavior of deep model` \cite{chen2019energy,chen2019hybrid} caused by ZSL, allowing the learning of all-sided attention information which might be useful for unseen identities, preventing the learning of biased attention information that only benefits to the seen identities.
\vspace{-0.5em}
\section{Proposed Approach}
In this section, we will first provide the formulation of the general attention mechanism in Sec. \ref{sec_attention}, then detail the proposed \emph{High-Order Attention} (HOA) module in Sec. \ref{sec_hoa}, finally show the overall framework of our \emph{Mixed High-Order Attention Network} (MHN) in Sec. \ref{sec_mhn}.
\subsection{Problem Formulation}\label{sec_attention}
Attention acts as a tool to bias the allocation of available resources towards the most informative parts of an input. In convolutional neural network (CNN), it is commonly used to reweight the convolutional response maps so as to highlight the important parts and suppress the uninformative ones, such as \emph{spatial attention} \cite{li2018diversity,li2018harmonious} and \emph{channel attention} \cite{hu2018squeeze,li2018harmonious}. We extend these two attention methods to a general case. Specifically, for a convolutional activation output, a 3D tensor $\mathcal{X}$, encoded by CNN and coming from the given input image. We have $\mathcal{X}\in{\mathbb{R}^{C\times H\times W}}$, where $C,H,W$ indicate the number of channel, height and width, \emph{resp}. As aforementioned, the goal of attention is to reweight the convolutional output, we thus formulate this process as:
\vspace{-0.6em}
\begin{equation}\label{eq_attention}
  \mathcal{Y}=\mathcal{A}(\mathcal{X})\odot{\mathcal{X}}
\end{equation}
where $\mathcal{A}(\mathcal{X})\in{\mathbb{R}^{C\times H\times W}}$ is the attention proposal output by a certain attention module, $\odot$ is the Hadamard Product (element-wise product). As $\mathcal{A}(\mathcal{X})$ serves as a reweighting term, the value of each element of $\mathcal{A}(\mathcal{X})$ should be in the interval $[0,1]$. Based on the above general formulation of attention, $\mathcal{A}(\mathcal{X})$ can take many different forms. For example, if $\mathcal{A}(\mathcal{X})=rep[M]|^{C}$ where $M\in{\mathbb{R}^{H\times W}}$ is a spatial mask and $rep[M]|^{C}$ means replicate this spatial mask $M$ along channel dimension by $C$ times, Eq. \ref{eq_attention} thus is the implementation of \emph{spatial attention}. And if $\mathcal{A}(\mathcal{X})=rep[V]|^{H,W}$ where $V\in{\mathbb{R}^{C}}$ is a scale vector and $rep[V]|^{H,W}$ means replicate this scale vector along height and width dimensions by $H$ and $W$ times \emph{resp}, Eq. \ref{eq_attention} thus is the implementation of \emph{channel attention}.

However, in \emph{spatial attention} or \emph{channel attention}, $\mathcal{A}(\mathcal{X})$ is coarse and unable to capture the high-order and complex interactions among parts, resulting in less discriminative attention proposals and failing in capturing the subtle differences among pedestrians. To this end, we dedicate to modeling $\mathcal{A}(\mathcal{X})$ with high-order statistics.
\subsection{High-Order Attention Module}\label{sec_hoa}
To model the complex and high-order interactions within attention, we first define a linear polynomial predictor on top of the high-order statistics of $\mathrm{\mathbf{x}}$, where $\mathrm{\mathbf{x}}\in{\mathbb{R}^{C}}$ denotes a local descriptor at a specific spatial location of  $\mathcal{X}$:
\vspace{-0.5em}
\begin{equation}\label{eq_def_a}
  a(\mathrm{\mathbf{x}})=\sum_{r=1}^{R}\langle\mathrm{\mathbf{w}}^{r},\otimes_{r}\mathrm{\mathbf{x}}\rangle
  \vspace{-0.5em}
\end{equation}
where $\langle\cdot,\cdot\rangle$ indicates the inner product of two same-sized tensors, $R$ is the number of order, $\otimes_{r}\mathrm{\mathbf{x}}$ is the r-th order outer-product of $\mathrm{\mathbf{x}}$ that comprises all the degree-r monomials in $\mathrm{\mathbf{x}}$, and $\mathrm{\mathbf{w}}^{r}$ is the r-th order tensor to be learned that contains the weights of degree-r variable combinations in $\mathrm{\mathbf{x}}$.

Considering $\mathrm{\mathbf{w}}^{r}$ with large $r$ will introduce excessive parameters and incur the problem of overfitting, we suppose that when $r>1$, $\mathrm{\mathbf{w}}^{r}$ can be approximated by $D^{r}$ rank-1 tensors by Tensor Decomposition \cite{kolda2009tensor}, i.e. $\mathrm{\mathbf{w}}^{r}=\sum^{D^{r}}_{d=1}\alpha^{r,d}\mathbf{u}_{1}^{r,d}\otimes\cdots\otimes\mathbf{u}_{r}^{r,d}$ when $r>1$, where $\mathbf{u}_{1}^{r,d}\in{\mathbb{R}^{C}},\ldots,\mathbf{u}_{r}^{r,d}\in{\mathbb{R}^{C}}$ are vectors, $\otimes$ is the outer-product, $\alpha^{r,d}$ is the weight for $d$-th rank-1 tensor. Then according to the tensor algebra, Eq. \ref{eq_def_a} can be reformulated as:
\vspace{-0.5em}
\begin{align}\label{eq_redef_a}
  a(\mathrm{\mathbf{x}})&=\langle\mathrm{\mathbf{w}}^{1},\mathrm{\mathbf{x}}\rangle+\sum_{r=2}^{R}\langle\sum_{d=1}^{D^{r}}\alpha^{r,d}\mathbf{u}_{1}^{r,d}\otimes\cdots\otimes\mathbf{u}_{r}^{r,d},\otimes_{r}\mathrm{\mathbf{x}}\rangle\nonumber\\
  &=\langle\mathrm{\mathbf{w}}^{1},\mathrm{\mathbf{x}}\rangle+\sum_{r=2}^{R}\sum_{d=1}^{D^{r}}\alpha^{r,d}\prod_{s=1}^{r}\langle\mathbf{u}_{s}^{r,d},\mathrm{\mathbf{x}}\rangle\nonumber\\
  &=\langle\mathrm{\mathbf{w}}^{1},\mathrm{\mathbf{x}}\rangle+\sum_{r=2}^{R}\langle\bm{\alpha}^{r},\mathbf{z}^{r}\rangle
\end{align}
where $\bm{\alpha}^{r}=[\alpha^{r,1},\cdots,\alpha^{r,D^{r}}]^{T}$ is the weight vector, $\mathbf{z}^{r}=[z^{r,1},\cdots,z^{r,D^{r}}]^{T}$ with $z^{r,d}=\prod_{s=1}^{r}\langle\mathbf{u}_{s}^{r,d},\mathrm{\mathbf{x}}\rangle$. For later convenience, Eq. \ref{eq_redef_a} can also be written as:
\vspace{-1em}
\begin{equation}\label{eq_conven redef_a}
  a(\mathrm{\mathbf{x}})=\mathbf{1}^{T}(\mathrm{\mathbf{w}}^{1}\odot\mathrm{\mathbf{x}})+\sum_{r=2}^{R}\mathbf{1}^{T}(\bm{\alpha}^{r}\odot\mathbf{z}^{r})
  \vspace{-0.5em}
\end{equation}
where $\odot$ is Hadamard Product and $\mathbf{1}^{T}$ is a row vector of ones. Then, to obtain a vector-like predictor $\mathbf{a}(\mathrm{\mathbf{x}})\in\mathbb{R}^{C}$, Eq. \ref{eq_conven redef_a} is generalized by introducing the auxiliary matrixes $\mathbf{P}^{r}$:
\vspace{-1.5em}
\begin{equation}\label{eq_conven_redef_a_vec}
  \mathbf{a}(\mathrm{\mathbf{x}})={\mathbf{P}^{1}}^{T}(\mathrm{\mathbf{w}}^{1}\odot\mathrm{\mathbf{x}})+\sum_{r=2}^{R}{\mathbf{P}^{r}}^{T}(\bm{\alpha}^{r}\odot\mathbf{z}^{r})
  \vspace{-0.3em}
\end{equation}
where $\mathbf{P}^{1}\in\mathbb{R}^{C\times C}$, $\mathbf{P}^{r}\in\mathbb{R}^{D^{r}\times C}$ with $r>1$. Since all $\mathbf{P}^{r},\mathrm{\mathbf{w}}^{1},\bm{\alpha}^{r}$ are parameters to be learned, for implementation convenience, we can integrate \{$\mathbf{P}^{1},\mathrm{\mathbf{w}}^{1}$\} into a new single matrix $\mathbf{\widehat{w}}^{1}\in{\mathbb{R}^{C\times C}}$ according to matrix algebra, and \{$\mathbf{P}^{r},\bm{\alpha}^{r}$\} into $\bm{\widehat{\alpha}}^{r}\in\mathbb{R}^{D^{r}\times C}$ (simple proof is in Supplementary file). Then Eq. \ref{eq_conven_redef_a_vec} can be expressed as:
\vspace{-0.5em}
\begin{equation}\label{eq_conven_redef_a_vec_inte}
  \mathbf{a}(\mathrm{\mathbf{x}})={~\mathbf{\widehat{w}}^{1}}^{T}\mathrm{\mathbf{x}}+\sum_{r=2}^{R}{~\bm{\widehat{\alpha}}^{r}}^{T}\mathrm{\mathbf{z}}^{r}
  \vspace{-0.5em}
\end{equation}

The above equation contains two terms, for clarity, we intend to formulate it into a more general case. Suppose $\mathbf{\widehat{w}}^{1}$ can be approximated by the multiplication of two matrixes $\widehat{\mathbf{v}}\in\mathbb{R}^{C\times D^{1}}$ and $\widehat{\bm{\alpha}}^{1}\in\mathbb{R}^{D^{1}\times C}$, i.e. $\mathbf{\widehat{w}}^{1}=\widehat{\mathbf{v}}\widehat{\bm{\alpha}}^{1}$. then Eq. \ref{eq_conven_redef_a_vec_inte} can be reformulated as:
\vspace{-1em}
\begin{equation}\label{eq_final_a}
  \mathbf{a}(\mathrm{\mathbf{x}})={~\widehat{\mathbf{\bm{\alpha}}}^{1}}^{T}(\widehat{\mathbf{v}}^{T}\mathbf{x})+\sum_{r=2}^{R}{~\bm{\widehat{\alpha}}^{r}}^{T}\mathrm{\mathbf{z}}^{r}=\sum_{r=1}^{R}{~\widehat{\bm{\alpha}}^{r}}^{T}\mathbf{z}^{r}
  \vspace{-0.7em}
\end{equation}
where $\mathbf{z}^{1}=\widehat{\mathbf{v}}^{T}\mathbf{x}$, and when $r>1$, $\mathbf{z}^{r}$ is the same as in Eq. \ref{eq_redef_a}. $\widehat{\bm{\alpha}}^{r}\in\mathbb{R}^{D^{r}\times C}$ are the trainable parameters.

In Eq.\ref{eq_final_a}, $\mathbf{a}(\mathrm{\mathbf{x}})$ is capable of modeling and using the high-order statistics of the local descriptor $\mathbf{x}$, thus, we can obtain the high-order vector-like attention `map' by performing Sigmoid function on Eq. \ref{eq_final_a}:
\vspace{-1em}
\begin{equation}\label{eq_sigmoid}
  A(\mathbf{x})=sigmoid\textbf{(}\mathbf{a}(\mathrm{\mathbf{x}})\textbf{)}=sigmoid(\sum_{r=1}^{R}{~\widehat{\bm{\alpha}}^{r}}^{T}\mathbf{z}^{r})
  \vspace{-0.7em}
\end{equation}
where $A(\mathbf{x})\in\mathbb{R}^{C}$ and the value of each element in $A(\mathbf{x})$ is in the interval $[0,1]$.

\textbf{Nonlinearity}: Moreover, in order to further improve the representation capacity of this high-order attention `map', inspired by the common design of CNN, we provide a variation of Eq.\ref{eq_sigmoid} by introducing nonlinearity as follows:
\vspace{-0.7em}
\begin{equation}\label{eq_nonlinearity}
  A(\mathbf{x})=sigmoid(\sum_{r=1}^{R}{~\widehat{\bm{\alpha}}^{r}}^{T}\sigma(\mathbf{z}^{r}))
  \vspace{-0.7em}
\end{equation}
where $\sigma$ denotes an arbitrary non-linear activation function, here, we use ReLU \cite{nair2010rectified} function. $A(\mathbf{x})$ in Eq.\ref{eq_nonlinearity} is finally employed as the required high-order attention `map' for the corresponding local descriptor $\mathbf{x}$. The experimental comparisons between Eq.\ref{eq_sigmoid} and Eq.\ref{eq_nonlinearity} are in Sec. \ref{sec_experiments}.

\textbf{Full module}: As aforementioned, $A(\mathbf{x})$ is defined on a local descriptor $\mathbf{x}$, to obtain $\mathcal{A}(\mathcal{X})$ which is defined on 3D tensor $\mathcal{X}$, we generalize Eq.\ref{eq_nonlinearity}. Specifically, we share the learnable weights in $A(\mathbf{x})$ among different spatial locations of $\mathcal{X}$ and let $\mathcal{A}(\mathcal{X})=\{A(\mathbf{x}_{(1,1)}),\cdots,A(\mathbf{x}_{(H,W)})\}$, where $\mathbf{x}_{(h,w)}$ indicates a local descriptor at spatial location point $(h,w)$ of $\mathcal{X}$. Employing this attention map $\mathcal{A}(\mathcal{X})$ in CNN has two benefits. (1) sharing weights among different spatial locations will not incur excessive parameters. (2) $\mathcal{A}(\mathcal{X})$ can be easily implemented by 1x1 convolution layers. After obtaining the high-order attention map $\mathcal{A}(\mathcal{X})$, our \emph{\textbf{High-Order Attention}} (HOA) module can be formulated in the same way as Eq. \ref{eq_attention}, i.e. $\mathcal{Y}=\mathcal{A}(\mathcal{X})\odot{\mathcal{X}}$.
\begin{figure}[t]
  \centering
  \includegraphics[width=0.9\linewidth]{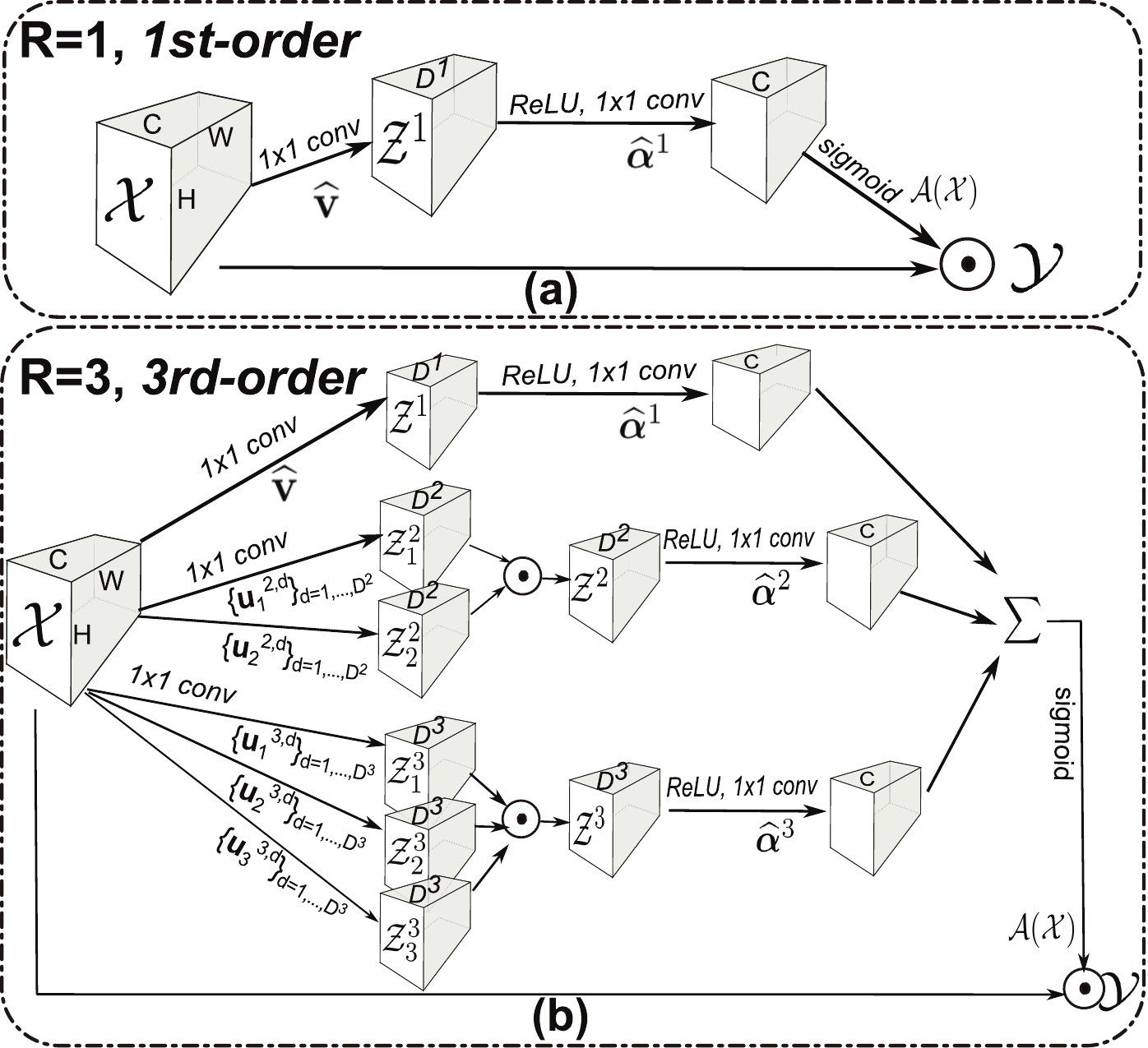}\\
  \caption{Illustration of \textbf{\emph{High-Order Attention}} (HOA) modules.}\label{fig_hoa}
  \vspace{-1.5em}
\end{figure}

\textbf{Implementation}: Since the learnable parameters are shared among spatial locations, all operations in $\mathcal{A}(\mathcal{X})$ can be implemented by Convolution. As illustrated in Fig. \ref{fig_hoa}.(a), when $R=1$, matrixes \{$\widehat{\mathbf{v}},\widehat{\bm{\alpha}}^{1}$\} are implemented by 1x1 convolution layers with $D^{1}$ and $C$ output channels, \emph{resp}. When $R>1,r>1$, we first employ $\{\mathbf{u}_{s}^{r,d}\}_{d=1,\cdots,D^{r}}$ as a set of $D^{r}$ 1x1 convolutional filters on $\mathcal{X}$ so as to produce a set of feature maps $\mathcal{Z}^{r}_{s}$ with channels $D^{r}$, then feature maps $\{\mathcal{Z}^{r}_{s}\}_{s=1,\cdots,r}$ are combined by element-wise product to obtain $\mathcal{Z}^{r}=\mathcal{Z}^{r}_{1}\odot\cdots\odot\mathcal{Z}^{r}_{r}$, where $\mathcal{Z}^{r}=\{\mathbf{z}^{r}\}$, and $\widehat{\bm{\alpha}}^{r}$ can also be implemented by 1x1 convolution layer. A toy example of HOA when $R=3$ is illustrated in Fig.\ref{fig_hoa}.(b).

\textbf{Remark}: The proposed HOA module can be easily implemented by the commonly used operations, such as 1x1 convolution and element-wise product/addition. Equipped by the powerful high-order predictor, the attention proposals could be more discriminative and is capable of modeling the complex and high-order relationships among parts. Moreover, the \emph{channel attention} module in \cite{hu2018squeeze,li2018harmonious} is called to be \emph{first-order} because (1) GAP layer only collects first-order statistics while neglecting richer higher-order ones, suffering from limited representation ability \cite{cimpoi2015deep} (2) fully-connected layers can be regarded as 1x1 convolution layers and thus the two cascaded fully-connected layers used in \emph{channel-attention} \cite{hu2018squeeze} are equivalent to our HOA module when $R=1$ (regardless of the spatial dimensions and see Fig.\ref{fig_hoa}.(a)). In summary, the full \emph{channel attention} module can only collect and utilize the first-order information, being insufficiently rich to capture the complex interactions and to produce the discriminative attention maps. And if without using GAP, the \emph{channel attention} module can be regarded as a special case of our HOA with $R=1$, further demonstrating it indeed is first-order.
\subsection{Mixed High-Order Attention Network}\label{sec_mhn}
\begin{figure}[t]
  \centering
  \includegraphics[width=1\linewidth]{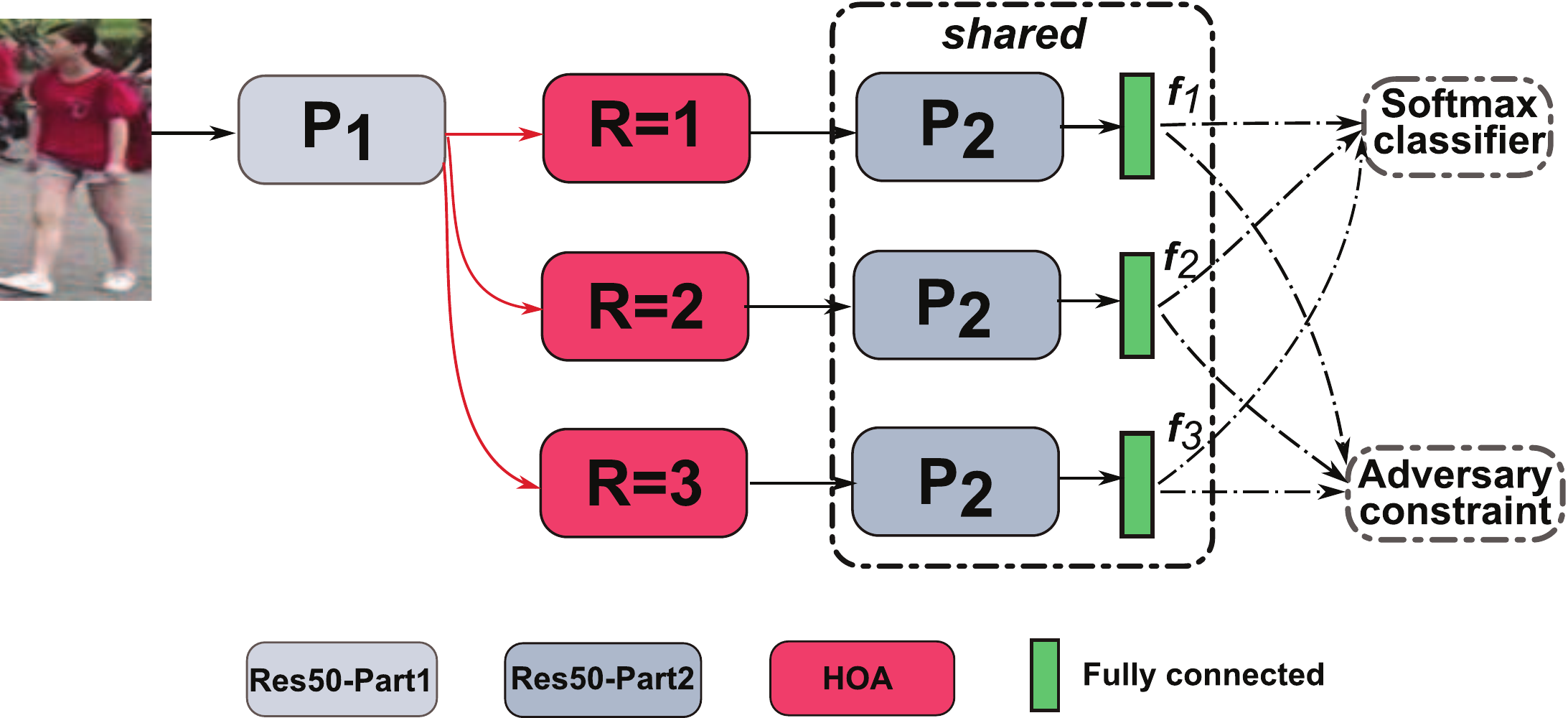}\\
  \caption{Illustration of \textbf{\emph{Mixed High-Order Attention Network}} (MHN). Our MHN is model-agnostic, it can be applied in both IDE \cite{zheng2016person} and PCB \cite{sun2018beyond} architectures, here for clarity, we take ResNet50 \cite{he2016deep} based IDE for example. The adversary constraint is used to regularize the order of HOA modules.}\label{fig_mhn}
  \vspace{-1.5em}
\end{figure}
Considering that Person ReID essentially pertains to \emph{zero-shot} learning (ZSL), where there is no intersection between training identities and testing identities, we should explicitly suppress the problem of `biased learning behavior of deep model' caused by \emph{zero-shot} settings \cite{chen2019energy}. Specifically, in ZSL, the deep model easily learn to focus on surface statistical regularities rather than more general abstract concepts, in other words, deep model will selectively learn the biased knowledge that are only useful to distinguish the seen identities, while ignore the knowledge that might be useful for unseen ones. Therefore, to correctly recognize the unseen identities, we propose \textbf{\emph{Mixed High-Order Attention Network}} (MHN) to utilize multiple HOA modules with different orders such that the diverse and complementary high-order information can be explicitly used, encouraging the richness of the learned features and preventing the learning of partial/biased visual information.

For a toy example as shown in Fig. \ref{fig_mhn}, the proposed MHN is constituted by several different HOA modules such that the diverse statistics of visual knowledge could be modeled and used. In particular, ResNet50 is first decomposed into two parts, i.e. P$_{1}$ (from \emph{conv1} to \emph{layer2}\footnote{Named in pytorch \cite{pytorch} manner.}) and P$_{2}$ (from \emph{layer3} to \emph{GAP}). P$_{1}$ is used to encode the given image from raw pixel space to mid-level feature space, P$_{2}$ is used to encode the attentional information to the high-level feature space where the data can be classified. HOA modules with different orders (e.g. \{$R=1,2,3$\}) are placed between P$_{1}$ and P$_{2}$ so as to produce the diverse high-order attention maps and intensify the richness within learned knowledge. Worthy of mention is that our MHN won't introduce excessive parameters since P$_{2}$ modules share the same weights across different attention streams.

However, simply employing multiple HOA modules with different orders won't lead the best performance of MHN, since one HOA module with higher order might collapse to a relatively lower order module due to `the partial/biased learning behavior of deep model'. Specifically, from Eq. \ref{eq_final_a}, one can observe that for a $k$-th order HOA module, $\mathbf{a}(\mathbf{x})$ also contains the $l$-th order sub-term (where $l<k$). In theory, HOA module with $R=k$ can capture and use the $k$-th order statistics of local descriptor $\mathbf{x}$, but in actual, especially in \emph{zero-shot} learning settings, due to the fact that the deep model will selectively learn surface statistical regularities that are the easiest ones to distinguish the seen classes \cite{chen2019energy}, the $k$-th order attention module might collapse to a lower-order counterpart as lower-order statistics are common and are much easier to collect than higher-order statistics. Therefore, these HOA modules with different $R$s actually collapse to some similar lower-order counterparts, and the wanted diverse higher-order attention information are not captured.
To this end, inspired by GAN \cite{goodfellow2014generative}, we introduce the adversary constraint for regularizing the order of HOA to be different, as shown in Fig. \ref{fig_mhn}, it can be formulated as:
\vspace{-0.8em}
\begin{small}
\begin{equation}\label{eq_adv}
  \max_{HOA|_{R=1}^{R=k}}\min_{F}(L_{adv})=\max_{HOA|_{R=1}^{R=k}}\min_{F}(\sum_{j,j^{'},j\neq j^{'}}^{k}\|F(f_{j})-F(f_{j^{'}})\|^{2}_{2})
  \vspace{-0.3em}
\end{equation}
\end{small}
where $HOA|_{R=1}^{R=k}$ means there are $k$ HOA modules (from first-order to $k$-th order) in MHN, $F$ indicates the encoding function parameterized by the adversary network which contains two fully-connected layers, $f_{j}$ is the feature representation vector learned from the corresponding HOA module with $R=j$. In Eq. \ref{eq_adv}, the adversary network $F$ tries to minimize the discrepancies among features $f_{j}$ while HOA modules try to maximize these discrepancies. After obtaining the Nash Equilibrium, the orders of HOA modules will be different with each other, since during the optimization of Eq.\ref{eq_adv}, P$_{2}$ shares across streams and the only differentiating parts in MHN are HOA modules, when maximizing the feature discrepancies, the only solution is to make the HOA modules have different orders and produce diverse attention knowledge. In other words, only diverse HOA modules will make $L_{adv}$ large. Thus the problem of order collapse can be suppressed.

Then, the overall objective function of MHN is as:
\vspace{-0.5em}
\begin{equation}\label{eq_mhn}
  \min(L_{ide})+\lambda(\max_{HOA|_{R=1}^{R=k}}\min_{F}(L_{adv}))
  \vspace{-0.5em}
\end{equation}
where $L_{ide}$ indicates the identity loss based on Softmax classifier, $\lambda$ is the coefficient.

\textbf{Remark}: From Eq.\ref{eq_mhn}, one can observe that we regularize the order/diversity of HOA modules by imposing constraint on the encoded feature vectors, instead of directly on the high-order attention maps, since these attention maps come from the complex high-order statistics and the definition of the order difference of HOA modules in the attention space is too hard to be artificially made. Thus, the order constraint is imposed on the feature vectors. Moreover, it seems that using Hinge loss based constraint instead of the adversary strategy to maximize the feature discrepancies is also feasible. However, we want to emphasize that in Hinge loss based function there is another margin-controller `m' which needs extra tuning, and the discrepancies between features that coming from different HOA modules will be heterogeneous, thus to determine the optimal margin `m', many redundant experiments must be executed. To this end, we employ the adversary constraint so as to allow the automatic learning of the optimal discrepancies.

By preventing the problem of order collapse, the HOA modules are explicitly regularized to model the wanted high-order attention distributions and thus can produce the discriminative and diverse attention maps which could be benefit for recognizing the unseen identities.
\vspace{-0.5em}
\section{Experiments}\label{sec_experiments}
\textbf{Datasets}: We use three popular benchmark datasets based on \emph{zero-shot} learning (ZSL) settings, i.e. \emph{\textbf{Market-1501}} \cite{zheng2015scalable}, \emph{\textbf{DukeMTMC-ReID}} \cite{ristani2016performance,zheng2017unlabeled} and \emph{\textbf{CUHK03-NP}} \cite{li2014deepreid,zhong2017re}. Market-1501 have 12,936 training images with 751 different identities. Gallery and query sets have 19,732 and 3,368 images respectively with another 750 identities. DukeMTMC-ReID includes 16,522 training images of 702 identities, 2,228 query and 17,661 gallery images of another 702 identities. CUHK03-NP is a new training-testing split protocol for CUHK03, it contains two subsets which provide labeled and detected (from a person detector) person images. The detected CUHK03 set includes 7,365 training images, 1,400 query images and 5,332 gallery images. The labeled set contains 7,368 training, 1,400 query and 5,328 gallery images respectively. The new protocol splits the training and testing sets into 767 and 700 identities.
\begin{table}[htbp]
  \centering
  \resizebox{1\linewidth}{!}{
    \begin{tabular}{c|c|cccc}
    \hline
    \multicolumn{2}{c}{} & \multicolumn{4}{c}{Market-1501 (\%)} \\
    \hline
    Methods & Ref   & R-1   & R-5   & R-10  & mAP \\
    \hline
     BoW+kissme \cite{zheng2015scalable}  & ICCV15  & 44.4  & 63.9  & 72.2 & 20.8 \\
     SVDNet \cite{sun2017svdnet}  & ICCV17  & 82.3  & -  & - & 62.1 \\
     DaRe(De)+RE \cite{wang2018resource} & CVPR18  & 89.0  & -  & - & 76.0 \\
     MLFN \cite{chang2018multi}& CVPR18  & 90.0  & -  & - & 74.3 \\
     KPM \cite{shen2018end}  & CVPR18  & 90.1  & 96.7  & 97.9 & 75.3 \\
     HA-CNN \cite{li2018harmonious}  & CVPR18  & 91.2  & -  & - & 75.7 \\
     DNN-CRF \cite{chen2018group}& CVPR18  & 93.5  & \textcolor[rgb]{0,0,1}{97.7}  & - & 81.6 \\
     PABR \cite{suh2018part} & ECCV18  & 91.7  & 96.9  & 98.1 & 79.6 \\
     PCB+RPP \cite{sun2018beyond}  & ECCV18  & 93.8  & 97.5  & 98.5 & 81.6 \\
     Mancs \cite{Wang_2018_ECCV}& ECCV18  & 93.1  & -  & - & 82.3 \\
      CASN+PCB \cite{zheng2018re} & CVPR19  &  \textcolor[rgb]{0,0,1}{94.4}  &   -    &   - & 82.8 \\
    \hline\hline
    IDE$^{*}$ \cite{zheng2016person}   &       & 89.0 & 95.7 & 97.3 & 73.9 \\
    \emph{MHN-6 (IDE)} &       & \emph{93.6}  & \emph{\textcolor[rgb]{0,0,1}{97.7}} & \emph{\textcolor[rgb]{0,0,1}{98.6}} & \emph{\textcolor[rgb]{0,0,1}{83.6}} \\
    \hline
    PCB$^{*}$ \cite{sun2018beyond}  &       & 93.1 & 97.5 & 98.5 & 78.6 \\
    \textbf{\emph{MHN-6 (PCB)}} &       & \textcolor[rgb]{1,0,0}{\emph{\textbf{95.1}}}  & \textcolor[rgb]{1,0,0}{\emph{\textbf{98.1}}}  & \textcolor[rgb]{1,0,0}{\emph{\textbf{98.9}}}  & \textcolor[rgb]{1,0,0}{\emph{\textbf{85.0}}} \\
    \hline
    \end{tabular}
    }
      \caption{Results comparisons over Market-1501 \cite{zheng2015scalable} under Single-Query settings. $^{*}$ indicates the re-implementation by our code. The best/second results are shown in red/blue, \emph{resp}.}
      \label{tab_Market}
\end{table}
\begin{table}[htbp]
\centering
\resizebox{1\linewidth}{!}{
\begin{tabular}{c|c|cccc}
\hline
    \multicolumn{2}{c}{} & \multicolumn{4}{c}{DukeMTMC-ReID (\%)} \\
    \hline
    Methods & Ref   & R-1   & R-5   & R-10  & mAP \\
    \hline
    BoW+kissme \cite{zheng2015scalable}  & ICCV15  & 25.1  & -  & - & 12.2 \\
    SVDNet \cite{sun2017svdnet}  & ICCV17  & 76.7  & -  & - & 56.8 \\
    DaRe(De)+RE \cite{wang2018resource} & CVPR18  & 80.2 & -  & - & 64.5 \\
    MLFN \cite{chang2018multi}& CVPR18  & 81.0  & -  & - & 62.8 \\
    KPM \cite{shen2018end}  & CVPR18  & 80.3  & 89.5  & 91.9 & 63.2 \\
    HA-CNN \cite{li2018harmonious}  & CVPR18  & 80.5  & -  & - & 63.8 \\
    DNN-CRF \cite{chen2018group}& CVPR18  & 84.9  & 92.3  & - & 69.5 \\
    PABR \cite{suh2018part} & ECCV18  & 84.4  & 92.2  & 93.8 & 69.3 \\
    PCB+RPP \cite{sun2018beyond}  & ECCV18  & 83.3  & -  & - & 69.2 \\
    Mancs \cite{Wang_2018_ECCV}& ECCV18  & 84.9  & -  & - & 71.8 \\
      CASN+PCB \cite{zheng2018re} & CVPR19  &  \textcolor[rgb]{0,0,1}{87.7}  &   -    &   - & 73.7 \\
    \hline\hline
    IDE$^{*}$ \cite{zheng2016person}   &       & 80.1  & 90.7  & 93.5  & 64.2 \\
    \emph{MHN-6 (IDE)} &       & \emph{87.5}  & \textcolor[rgb]{0,0,1}{\emph{93.8}}  & \textcolor[rgb]{0,0,1}{\emph{95.6}}  & \textcolor[rgb]{0,0,1}{\emph{75.2}} \\
    \hline
    PCB$^{*}$ \cite{sun2018beyond}  &       & 83.9  & 91.8  & 94.4  & 69.7 \\
    \textbf{\emph{MHN-6 (PCB)}} &       & \textcolor[rgb]{1,0,0}{\emph{\textbf{89.1}}}  & \textcolor[rgb]{1,0,0}{\emph{\textbf{94.6}}}  & \textcolor[rgb]{1,0,0}{\emph{\textbf{96.2}}}  & \textcolor[rgb]{1,0,0}{\emph{\textbf{77.2}}} \\
    \hline
\end{tabular}}
  \caption{Results comparisons over DuckMTMC-ReID \cite{ristani2016performance,zheng2017unlabeled}. $^{*}$ indicates the re-implementation by our code. The best/second results are shown in red/blue, \emph{resp}.}\label{tab_Duke}
\end{table}
\begin{table}[htbp]
\centering
\resizebox{1\linewidth}{!}{
\begin{tabular}{c|c|cc|cc}
\hline
          &       & \multicolumn{4}{c}{CUHK03-NP (\%)} \\
          \hline
    \multirow{2}[0]{*}{Methods} & \multirow{2}[0]{*}{Ref} & \multicolumn{2}{c|}{Labeled} & \multicolumn{2}{c}{Detected} \\
          &       & R-1   & mAP   & R-1   & mAP \\
\hline
    BoW+XQDA \cite{zheng2015scalable}  & ICCV15  & 7.9  & 7.3  & 6.4 & 6.4 \\
    SVDNet \cite{sun2017svdnet}  & ICCV17  & - & -  & 41.5 & 37.3 \\
    DaRe(De)+RE \cite{wang2018resource} & CVPR18  & 66.1  & 61.6  & 63.3 & 59.0 \\
    MLFN \cite{chang2018multi}& CVPR18  & 54.7  & 49.2  & 52.8 & 47.8 \\
    HA-CNN \cite{li2018harmonious}  & CVPR18  & 44.4  & 41.0  & 41.7 & 38.6 \\
    PCB+RPP \cite{sun2018beyond}  & ECCV18  & -  & -  & 63.7 & 57.5 \\
    Mancs \cite{Wang_2018_ECCV}& ECCV18  & 69.0  & 63.9  & 65.5 & 60.5 \\
      CASN+PCB \cite{zheng2018re} & CVPR19  &  \textcolor[rgb]{0,0,1}{73.7}  &  \textcolor[rgb]{0,0,1}{68.0}    &  \textcolor[rgb]{0,0,1}{71.5} & \textcolor[rgb]{0,0,1}{64.4} \\
          \hline\hline
    IDE$^{*}$ \cite{zheng2016person}   &       & 52.9  & 48.5 & 50.4  & 46.3 \\
    \emph{MHN-6 (IDE)} &       & \emph{69.7}  & \emph{65.1}  & \emph{67.0}    & \emph{61.2} \\
    \hline
    PCB$^{*}$ \cite{sun2018beyond}  &       & 61.9  & 56.8  & 60.6  & 54.4 \\
    \textbf{\emph{MHN-6 (PCB)}} &       & \textcolor[rgb]{1,0,0}{\textbf{\emph{77.2}}}  & \textcolor[rgb]{1,0,0}{\emph{\textbf{72.4}}}  & \textcolor[rgb]{1,0,0}{\emph{\textbf{71.7}}}  & \textcolor[rgb]{1,0,0}{\emph{\textbf{65.4}}} \\
    \hline
    \end{tabular}%
    }
      \caption{Results comparisons over CUHK03-NP \cite{li2014deepreid,zhong2017re}. $^{*}$ indicates the re-implementation by our code. The best/second results are shown in red/blue, \emph{resp}.}
      \vspace{-0.5em}
  \label{tab_CUHK}%
\end{table}%

\textbf{Implementation}: The proposed MHN is applied on both ResNet50-based IDE \cite{zheng2016person} and PCB \cite{sun2018beyond} architectures. For both architectures, we adopt the SGD optimizer with a momentum factor of 0.9, set the start learning rate to be 0.01 for backbone CNN and ten times learning rate for the new added layers, and a total of 70 epochs with the learning rate decreased by a factor of 10 each 20 epochs. The dimension of feature $f_{j}$ is 256 and the two FC layers in $F$ have 128, 128 neurons \emph{resp}, we set all $D^{r}|_{r=1}^{R}$ to be 64. 
For IDE, the images are resized to 288x144. For PCB, the images are resized to 336x168. 
We set the batch size to 32 in all experiments and use one 1080Ti GPU. MHN is implemented by Pytorch \cite{pytorch} and modified from the public code\cite{layumi}, random erasing\cite{zhong2017random} is also applied.
\begin{table*}[htbp]
  \centering
  \resizebox{0.8\linewidth}{!}{
    \begin{tabular}{c|cc|cc|cccc|cccc}
    \hline
          & \multicolumn{4}{c|}{CUHK03-NP \cite{li2014deepreid,zhong2017re}} & \multicolumn{4}{c|}{\multirow{2}[4]{*}{DukeMTMC-ReID \cite{ristani2016performance,zheng2017unlabeled}}} & \multicolumn{4}{c}{\multirow{2}[4]{*}{Market-1501 \cite{zheng2015scalable}}}\\
\cline{2-5}    \multirow{2}[4]{*}{Methods} & \multicolumn{2}{c|}{Labeled} & \multicolumn{2}{c|}{Detected} & \multicolumn{4}{c|}{}         & \multicolumn{4}{c}{} \\
\cline{2-13}          & R-1   & mAP   & R-1   & mAP   & R-1   & R-5   & R-10  & mAP   & R-1   & R-5   & R-10  & mAP\\
    \hline
    IDE$^{*}$ \cite{zheng2016person}   &  52.9  & 48.5 & 50.4  & 46.3  &  80.1  & 90.7  & 93.5  & 64.2   &  89.0 & 95.7 & 97.3 & 73.9 \\
    IDE$^{*}$+era &  61.4   &  55.71   &  56.9  &   51.3  &   83.6  &  92.1   &  94.3   &  67.4   &  90.3  &   96.5  &   97.6  & 75.9\\
    MHN-2 (IDE) &  65.9  &   59.1  &  60.9   &  54.8   &  84.5   &  92.6   &  94.7   &  68.9   &   90.6   &   96.1   &   97.6  & 76.1 \\
    MHN-4 (IDE) &  67.4  &  60.3  &   62.7  &   55.8  &   86.3   &  93.1   &  95.6     &  72.4   &  91.8   &  97.6   &  98.5   & 80.1 \\
    MHN-6 (IDE) &   \textbf{69.7}  &   \textbf{65.1}  &   \textbf{67.0}  &  \textbf{61.2}   &   \textbf{87.5}  &   \textbf{93.8}  &  \textbf{ 95.6}   &  \textbf{75.2}   &  \textbf{93.6}  &  \textbf{97.7}   &  \textbf{98.6}   &\textbf{83.6}\\
    \hline
    PCB$^{*}$ \cite{sun2018beyond}   &  61.9  & 56.8  & 60.6  & 54.4    &   83.9  & 91.8  & 94.4  & 69.7   &  93.1 & 97.5 & 98.5 & 78.6 \\
    PCB$^{*}$+era &  57.4   &  52.5   &  54.3  &   49.9  &   83.4  &  91.5   &  94.3   &  68.2   &  91.9  &   97.4  &   98.4  & 76.8\\
    MHN-2 (PCB) &  71.2   &  66.3   &  67.9  &   61.9  &   86.9  &  93.3   &  95.3   &  73.5   &  94.0  &   97.8  &   98.5  & 82.5\\
    MHN-4 (PCB) &   75.1  &  70.6   &   71.6  &   \textbf{66.1}  &  88.7   &  94.4   &  95.9   &   76.8  &  94.5   &  98.0   &  98.6   & 84.2 \\
    MHN-6 (PCB) &  \textbf{77.2}  & \textbf{72.4} & \textbf{71.7}  & 65.4  &   \textbf{89.1}  & \textbf{94.6}& \textbf{96.2}  & \textbf{77.2}    &  \textbf{95.1}  & \textbf{98.1}  & \textbf{98.9}  & \textbf{85.0}  \\
    \hline
    \end{tabular}%
    }
    \vspace{0.1em}
      \caption{Effect (\%) of attention modules. $^{*}$ indicates the re-implementation and `era' means random erasing.}
      \vspace{-1em}
  \label{tab_mhn}%
\end{table*}%
\textbf{Notation}: We use `MHN-$k$' to denote that in MHN there are $k$ HOA modules with orders $R=\{1,\cdots,k\}$ \emph{resp}, and `MHN-$k$ (IDE/PCB)' to denote using IDE/PCB architectures, \emph{resp}.

\textbf{Evaluation}: In testing, the feature representations $f_{j},j\in\{1,\cdots,k\}$ are concatenated after L2 normalization. Then, the metrics of cumulative matching characteristic (CMC) and mean Average Precision (mAP) are used for evaluation. \textbf{\emph{No re-ranking tricks are adopted}}.\vspace{-0.6em}
\subsection{Comparison with State-of-the-Art Methods}\vspace{-0.6em}
In order to highlight the significance of the proposed MHN for person ReID task, we compare it with some recent remarkable works, including methods of alignment \cite{shen2018end,suh2018part,zheng2018re,sun2018beyond}, deep supervision \cite{wang2018resource}, architectures \cite{zheng2016person,sun2018beyond}, attention \cite{li2018harmonious,zheng2018re,Wang_2018_ECCV} and others \cite{sun2017svdnet,chen2018group,chang2018multi}, over the popular used benchmarks Market-1501, DukeMTMC-ReID and CUHK03-NP. For fair comparison, we re-implement the baseline models, i.e. ResNet50-based IDE and PCB, with the same training configurations as ours. MHN is then applied over both IDE and PCB architectures. The comparison results are listed in Tab. \ref{tab_Market}, Tab. \ref{tab_Duke} and Tab. \ref{tab_CUHK}. From these tables, one can observe that by explicitly intensify the discrimination and diversity within the deep embedding via high-order attention modules, our MHN-$6$ can significantly improve the performances over both the baseline methods IDE and PCB (e.g. comparing with PCB, MHN-$6$ (PCB) has $2\%/6.4\%$ improvements of R-$1$/mAP on Market and $5.2\%/7.5\%$ improvements of R-$1$/mAP on DukeMTMC), demonstrating the effectiveness of our high-order attention idea. And our MHN-$6$ (PCB) achieves the new SOTA performances on all these three benchmarks, showing the superiority of our method.
\vspace{-0.5em}
\subsection{Component Analysis}
\textbf{Effect of MHN}: We conduct quantitative comparisons on MHN as in Tab. \ref{tab_mhn}. From this table, one can observe that the proposed MHN can significantly improve the performances of person ReID task over both IDE and PCB baseline architectures. Specifically, comparing MHN-$2$(IDE/PCB) with IDE/PCB, we can see that using higher-order attention information indeed encourage the discrimination of the learned embedding. Moreover, the performances will further increase with the number of HOA modules, e.g. on CUHK03-NP Labeled dataset, applying MHN on PCB, when increasing the number of HOA modules from $2$ to $6$ the performance of R-$1$ will be increased from $71.2\%$ to $77.2\%$, the same phenomenon can be observed in other datasets and architecture. This phenomenon also shows that employing multiple HOA modules is benefit for modeling diverse and discriminative information for recognizing the unseen identities, and MHN-$6$ outperforms all the baseline models by a large margin over all the three benchmarks, demonstrating the effectiveness of our method. However, when further increase the number of HOA modules, e.g. $k=8$, the performance improvements are few, thus we don't report it here. 
\begin{table}[t!]
  \centering
  \resizebox{1\linewidth}{!}{
    \begin{tabular}{c|cc|cc}
    \hline
    \multirow{2}[3]{*}{Methods} & \multicolumn{2}{c|}{DukeMTMC-ReID} &\multicolumn{2}{c}{Market-1501}\\
\cline{2-5}           & R-1   & mAP   & R-1    & mAP\\
    \hline
    IDE$^{*}$ \cite{zheng2016person}    &  80.1  & 64.2  & 89.0  & 73.9   \\
    MHN-6 (IDE) w/o $L_{adv}$ &    85.5   &  70.8   &  91.8     &  80.0   \\
    MHN-6 (IDE) &    \textbf{87.5}  &   \textbf{75.2}  &  \textbf{ 93.6}   &  \textbf{83.6}  \\
    \hline
    PCB$^{*}$ \cite{sun2018beyond}   &   83.9  & 69.7  & 93.1  & 78.6  \\
    MHN-6 (PCB) w/o $L_{adv}$   &  87.7   &  75.4   &  93.9   &   83.2   \\
    MHN-6 (PCB) &  \textbf{89.1}  & \textbf{77.1}& \textbf{95.1}  & \textbf{85.0} \\
    \hline
    \end{tabular}%
    }
      \caption{Effect (\%) of adversary constraint. $^{*}$ indicates the re-implementation by our code.}
      \vspace{-1.5em}
  \label{tab_adv}%
\end{table}%
\begin{table}[t!]
  \centering
  \resizebox{1\linewidth}{!}{
    \begin{tabular}{c|cc|cc}
    \hline
    \multirow{2}[3]{*}{Methods} & \multicolumn{2}{c|}{DukeMTMC-ReID} &\multicolumn{2}{c}{Market-1501}\\
\cline{2-5}           & R-1   & mAP   & R-1    & mAP\\
    \hline
    MHN-6 (IDE) w/o $nonli$ &    87.1   &  74.9   &  93.3     &  83.1   \\
    MHN-6 (IDE) &    \textbf{87.5}  &   \textbf{75.2}  &  \textbf{ 93.6}   &  \textbf{83.6}  \\
    \hline
    MHN-6 (PCB) w/o $nonli$   &  88.7   &  76.8   &  95.0   &   84.5   \\
    MHN-6 (PCB) &  \textbf{89.1}  & \textbf{77.1}& \textbf{95.1}  & \textbf{85.0} \\
    \hline
    \end{tabular}%
    }
      \caption{Effect (\%) of nonlinearity.}
      \vspace{-1.5em}
  \label{tab_nonl}%
\end{table}%

\textbf{Effect of Adversary Constraint}: From Tab. \ref{tab_adv}, when comparing \{MHN-$6$ (IDE) w/o $L_{adv}$\} with \{IDE\} and comparing \{MHN-$6$ (PCB) w/o $L_{adv}$\} with \{PCB\}, one can observe that on both DukeMTMC and Market datasets the performances of R-$1$ and mAP can be improved by simply employing multiple HOA modules without any regularizing constraint, showing that using higher-order attention information will indeed increase the discrimination of the learned knowledge in ZSL settings. However, as mentioned in Sec. \ref{sec_mhn}, the task of person ReID pertains to \emph{zero-shot} settings, the problem of `partial/biased learning behavior of deep model' will incur the problem of order collapse of HOA modules, i.e. the deep model will partially model the easy and lower-order information regardless the theoretical capacity of HOA module. Therefore, we introduce the adversary constraint to explicitly prevent the problem of order collapse. After equipping with $L_{adv}$, MHN-$6$(IDE/PCB) can further improve the performances over both the benchmarks, demonstrating the effectiveness of $L_{adv}$ and implying that explicitly learning diverse high-order attention information is essential for recognizing the \emph{unseen} identities.

\textbf{Effect of Nonlinearity}: The nonlinearity comparisons are listed in Tab. \ref{tab_nonl}, from this table, one can observe that by adding nonlinearity into the high-order attention modules, the performances can be further improved.

\begin{table}[t!]
  \centering
  \resizebox{0.9\linewidth}{!}{
    \begin{tabular}{c|cc|cc}
    \hline
    \multirow{2}[3]{*}{Methods} & \multicolumn{2}{c|}{DukeMTMC-ReID} &\multicolumn{2}{c}{Market-1501}\\
\cline{2-5}           & R-1   & mAP   & R-1    & mAP\\
    \hline
    IDE$^{*}$ \cite{zheng2016person}    &  80.1  & 64.2  & 89.0  & 73.9   \\
    SENet50$^{*}$ \cite{hu2018squeeze} &81.2 & 64.8 & 90.0 & 75.6 \\
    HA-CNN \cite{li2018harmonious} & 80.5 & 63.8 & 91.2 & 75.7 \\
    SpaAtt+Q$^{*}$ \cite{li2018diversity} &84.7&69.6 & 91.6 & 77.4\\
    CASN+IDE \cite{zheng2018re} & 84.5 & 67.0 & 92.0 & 78.0 \\
    MHN-6 (IDE) &    \textbf{87.5}  &   \textbf{75.2}  &  \textbf{ 93.6}   &  \textbf{83.6}  \\
    \hline
    \end{tabular}%
    }
      \caption{Comparison to other attention methods (\%). $^{*}$ indicates our reproducing.}
      \vspace{-1.5em}
  \label{tab_atten_comp}%
\end{table}%
\textbf{Comparison to other attention methods}: To demonstrate the effectiveness of our idea of high-order attention, we compare with some other attention methods as in Tab. \ref{tab_atten_comp}. Specifically, our MHN-$6$(IDE) outperforms both the \emph{spatial} and \emph{channel} attention methods, i.e. HA-CNN \cite{li2018harmonious} and SENet50 \footnote{We fine-tune the pre-trained SENet50 released at \url{https://github.com/moskomule/senet.pytorch}.} \cite{hu2018squeeze}, showing the superiority of high-order attention model to these coarse/first-order attention methods. Moreover, although \{SpaAtt+Q\} \cite{li2018diversity} employs multiple diverse attention modules like MHN to enhance the richness of attention information, the used attention method is \emph{spatial attention} which is coarse and insufficiently rich to capture the complex and high-order interactions of parts, failing in producing more discriminative attention proposals and thus performing worse than MHN-$6$(IDE). \{CASN+IDE\} \cite{zheng2018re} regularizes the attention maps of the paired images belonging to the same identity to be similar and indeed improves the results, but it still performs worse than MHN-$6$(IDE) since the consistence constraint for attention maps is only based on the the coarse \emph{spatial attention} maps.

In summary, because of the ability of modeling and using complex and high-order information, the proposed MHN can significantly surpass all the listed coarse/first-order attention methods as shown in Tab. \ref{tab_atten_comp}.

\textbf{Ablation study on the configurations of P$_{1}$ \& P$_{2}$}: As mentioned in Sec. \ref{sec_mhn}, the HOA modules are placed between P$_{1}$ and P$_{2}$, to investigate the effect of the placed position of HOA modules, we conduct experiments as in Tab. \ref{tab_p1p2}. One can observe that placing HOA modules after `\emph{layer2}' (i.e. using the configuration \textcircled{2}) performs the best since when placing it at the relatively lower layer (i.e. using the configuration \textcircled{1}) the knowledge input to HOA module is more relevant to the low-level texture information and contains much noise, while placing it at relatively higher layer (i.e. using the configuration \textcircled{3}), some useful knowledge for recognizing the unseen identities might be already lost during the forward propagation of information as a result of partial/biased learning behavior. To this end, we use the configuration \textcircled{2} for both IDE and PCB architectures throughout the experiments.
\begin{table}[t!]
  \centering
  \resizebox{1\linewidth}{!}{
    \begin{tabular}{c|cc}
   \hline
    \multirow{2}[3]{*}{Methods} & \multicolumn{2}{c}{Market-1501} \\
\cline{2-3}          & R-1   & mAP \\
    \hline
    \textcircled{1}:P$_{1}$=\{\emph{conv1}$\sim$\emph{layer1}\},P$_{2}$=\{\emph{layer2}$\sim$\emph{GAP}\}&  92.2 & 81.8 \\
    \textcircled{2}:P$_{1}$=\{\emph{conv1}$\sim$\emph{layer2}\},P$_{2}$=\{\emph{layer3}$\sim$\emph{GAP}\}&  \textbf{93.6} & \textbf{83.6} \\
    \textcircled{3}:P$_{1}$=\{\emph{conv1}$\sim$\emph{layer3}\},P$_{2}$=\{\emph{layer4}$\sim$\emph{GAP}\}&  92.7 & 82.1 \\
    \hline
    \end{tabular}%
    }\vspace{0.1em}
      \caption{Ablation study on the configurations of P$_{1}$ and P$_{2}$. All the layer names are shown in Pytorch manner. Here, for convenience we conduct experiments with MHN-$6$ (IDE) and test three configurations, i.e. \textcircled{1},\textcircled{2} and \textcircled{3}.}
      \vspace{-0.5em}
  \label{tab_p1p2}%
\end{table}%

\textbf{Model size}: We compare the model size as in Tab. \ref{tab_complexity}, from this table, one can observe that the parameter number of our MHN increases with the order. While comparing with SENet50 \cite{hu2018squeeze}, the total parameter number of each MHN is not so much, and in terms of the performance, each MHN can outperform SENet50, showing that our MHN is indeed `light and sweet'.
\begin{table}
  \centering
  \resizebox{1\linewidth}{!}{
  \begin{tabular}{c|c|c|c}
     \hline
     Models & PN (million) & Depth & R-1 (on Market) \\
     \hline
     IDE \cite{zheng2016person}  & 24.2  & 50 & 89.0\%\\
     SENet50 \cite{hu2018squeeze} & 27.4 & 50 & 90.0\%\\
     MHN-$2$ (IDE) & 24.4 & 50 & 90.6\%\\
     MHN-$4$ (IDE) & 25.2 & 50 & 91.8\%\\
     MHN-$6$ (IDE) & 26.8 & 50 & 93.6\%\\
     \hline
   \end{tabular}
   }
  \caption{Model size comparisons. PN means Parameter Number.}\label{tab_complexity}
  \vspace{-1.5em}
\end{table}
\vspace{-0.5em}
\section{Conclusion}
\vspace{-0.8em}
In this paper, we first propose the High-Order Attention (HOA) module so as to increase the discrimination of attention proposals by modeling and using the complex and high-order statistics of parts. Then, considering the fact that the person-ReID task pertains to \emph{zero-shot} learning where the deep model will easily learn the biased knowledge, we propose the Mixed High-Order Attention Network (MHN) to utilize the HOA modules at different orders, preventing the learning of partial/biased visual information that only benefit to the seen identities. The adversary constraint is further introduced to prevent the problem of order collapse of HOA module. And Extensive experiments have been conducted over three popular benchmarks to validate the necessity and effectiveness of our method.

\small{\textbf{Acknowledgments}: This work was partially supported by the National Natural Science Foundation of China under Grant Nos. 61871052, 61573068, 61471048, and BUPT Excellent Ph.D. Students Foundation CX2019307.}
{\small
\bibliographystyle{ieee_fullname}
\bibliography{egbib}
}

\end{document}